\documentclass[conference]{IEEEtran}
\IEEEoverridecommandlockouts
\usepackage{cite}
\usepackage{amsmath,amssymb,amsfonts}
\usepackage{algorithmic}
\usepackage{graphicx}
\usepackage{textcomp}
\usepackage{xcolor}
\usepackage{soul} % for \hl highlighting
\usepackage{afterpage}         % for scheduling floats on the next page
\usepackage{dblfloatfix}
\usepackage{xurl}
\sethlcolor{yellow}

\def\BibTeX{{\rm B\kern-.05em{\sc i\kern-.025em b}\kern-.08em
    T\kern-.1667em\lower.7ex\hbox{E}\kern-.125emX}}

\begin{document}

\title{ASCII Art Turns LLMs into VLA Controllers}

% \author{Hidden}

\author{%
Yitao Jiang\textsuperscript{1},
Roy Xing\textsuperscript{1},
Luyang Zhao\textsuperscript{2},
Brian Plancher\textsuperscript{1},
Muhao Chen\textsuperscript{3},
Devin Balkcom\textsuperscript{1}%
}

\maketitle
\begingroup
\renewcommand\thefootnote{\arabic{footnote}}
\setcounter{footnote}{0}
\footnotetext[1]{Dept. of Computer Science, Dartmouth College, Hanover, NH, USA. Email: \{yitao.jiang.gr, roy.xing.gr, brian.k.plancher, devin.balkcom\}@dartmouth.edu}
\footnotetext[2]{Dept. of Electrical and Computer Engineering, Clemson University, Clemson, SC, USA. Email: luyangz@clemson.edu}
\footnotetext[3]{Dept. of Mechanical and Aerospace Engineering, University of Houston, Houston, TX, USA. Email: muhaochen@uh.edu}
\endgroup

\begin{abstract}
Vision--Language--Action (VLA) controllers are often built by extending vision--language models (VLMs) with action supervision, relying on multimodal backbones with large data and compute requirements. We demonstrate that a text-only large language model (LLM) can be adapted into a VLA-style controller when visual observations are rendered into a text input using an ASCII representation. This ASCII-as-vision interface enables existing training and deployment stacks for LLMs to efficiently condition on visual state, follow natural-language instructions, and produce constrained, executable actions. We fine-tune and compare multiple LLMs and VLMs across model families and scales, using both expert demonstrations from a planning-based teacher, as well as DAgger for iterative improvement. In a 2D manipulation benchmark, in both simulation and on a physical manipulator, the resulting controllers can identify task-relevant entities and plan feasible action sequences. Our results suggest that ASCII rendering can serve as a lightweight, interpretable modality bridge from images to text, complementing conventional VLA pipelines, and opening directions for VLA research with text-only backbones.
\end{abstract}

% \begin{IEEEkeywords}
% Vision Language Action, Vision Language Models, Large Language Models, ASCII Art, Robot Manipulation
% \end{IEEEkeywords}

\section{Introduction}

\subsection{Problem \& Motivation}
VLA models offer an appealing paradigm for robot control: by mapping visual observations and natural-language instructions directly to executable actions, a single model can unify perception, instruction understanding, planning, and control~\cite{kawaharazuka_vision-language-action_2025}. While modular pipelines with explicit optimization and autonomy stacks remain highly effective in practice (e.g.,~\cite{kuindersma_optimization-based_2016,tranzatto_cerberus_2022}), VLA aims to reduce engineering overhead and improve adaptability across tasks.

\begin{figure}[h]
  \centering
  \includegraphics[width=\columnwidth]{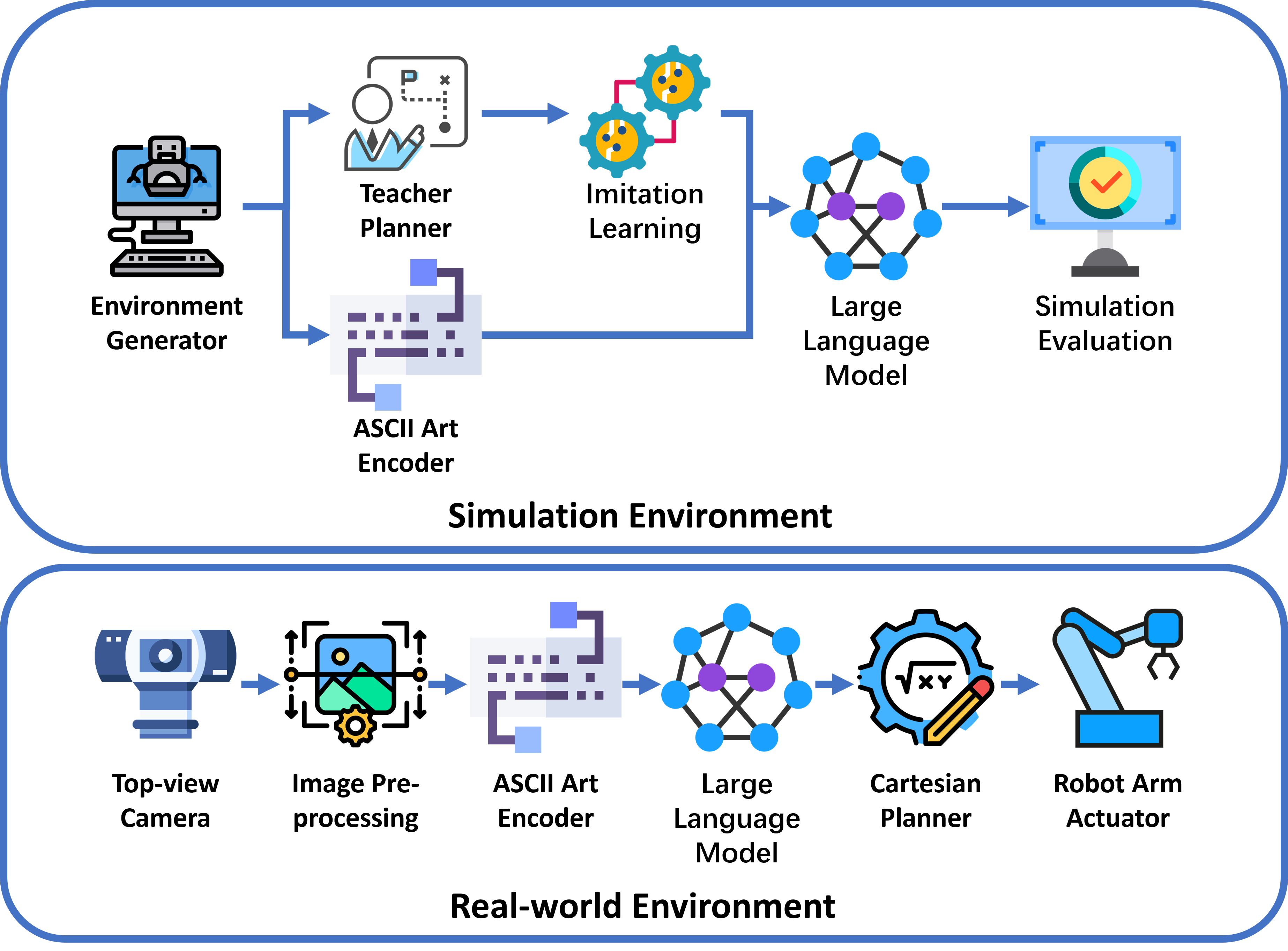}
  \caption{End-to-end pipeline showcasing training and testing in simulation (top), and real-robot deployment (bottom). \emph{Top}: a procedural environment generator produces scenes that are rendered to RGB, converted to a fixed-size ASCII art representation, and paired with expert actions from a planning-based teacher to train an LLM controller via imitation learning; the resulting policy is evaluated in closed loop. \emph{Bottom}: an overhead camera stream is pre-processed and encoded into ASCII art, the LLM outputs a command, and a Cartesian planner converts it to robot arm actions for execution.}
  \label{fig:pipeline}
  \vspace{-15pt}
\end{figure}

Most VLA systems build on VLMs. In a typical pipeline, a VLM interprets an image conditioned on language and is further trained through demonstrations to output robot actions~\cite{zitkovich_rt-2_2023, driess_palm-e_2023, kim_openvla_2024, ghosh_octo_2024}. Many such approaches rely on large-scale human teleoperation to collect sufficient high-quality robot (manipulation) demonstration data~\cite{brohan_rt-1_2023, kim_openvla_2024, collaboration_open_2025}. Simulation can reduce the human cost, but suffers from sim-to-real discrepancies that make transfer to real hardware difficult~\cite{goldberg_good_2025}. At the same time, training and deploying VLM-based controllers is compute-intensive, raising the barrier to exploring VLA systems and iterating on new ideas~\cite{shukor_smolvla_2025, reuss_flower_2025}. These constraints and limitations motivate the research question for this work: \emph{Can we build a VLA model using just LLMs without direct image inputs or multimodal training pipelines?}

\subsection{Key Idea: ASCII-as-Vision}

ASCII art has two practical advantages as a visual representation. First, it is compatible with text-only LLMs, avoiding additional vision-modality adaptation to create a VLM. In fact, ASCII art has been studied as a diagnostic medium for evaluating models' ability to perceive visual semantics embedded in text strings~\cite{jia_asciieval_2025,luo_asciibench_2025}. Second, it is human-readable, making observations and decisions easier to log and debug. Importantly, we do not claim that ASCII art is a more information-efficient representation than standard visual embeddings, as it inevitably discards some visual nuance in the conversion. Rather, we view ASCII art as a direct representation to utilize widely available LLMs and overcome data gathering constraints. Our results also suggest that in the 2D manipulation case, text-only models even perform better than the equivalent visual models.

As such, in this work, we create an ASCII art conditioned VLA controller to produce actions for manipulation tasks, including goal reaching and pick-and-place in the presence of obstacles. We train in a 2D manipulation environment with procedurally generated scenes. To bootstrap learning, we generate demonstrations from an expert policy based on traditional motion planning methods, such as A* \cite{4082128}, and we iteratively improve the controller using DAgger~\cite{ross_reduction_2011}. Our setup also includes task reachability assessments. In scenes where no feasible movement plan is possible due to obstacles, the controller recognizes as such and reports the infeasibility. We validate our approach on a 7 DOF manipulator and conduct tasks in the real world. Fig.~\ref{fig:pipeline} summarizes our end-to-end pipeline, and Fig.~\ref{fig:real_experiment} shows a representative real-robot rollout together with the associated image pre-processing and viewpoint correction, and the final ASCII encoding in the bottom right.
%
% \subsection{Contributions}
In summary, this work makes three main contributions:
\begin{itemize}
  \item We propose and evaluate a novel ASCII-art-based VLA, which uses LLMs to condition on visual state without native image inputs. The resulting controller is also able to evaluate reachability in the task setting.
  \item We build a reproducible training pipeline with an environment generator, a motion planning teacher policy, and DAgger-style iterative data collection.
  \item We benchmark multiple text-only LLM and vision-language model families and scales in simulation, and demonstrate successful sim-to-real transfer.
\end{itemize}

\section{Related Work}
\subsection{VLM and Multimodal Pretraining for Robotics}
Recent robot manipulation methods have been driven by scaling multimodal pretraining and imitation learning with large, diverse robot datasets. Early work, such as RT-1, demonstrated that transformer policies trained on real-robot data can exhibit improved generalization with scale \cite{brohan_rt-1_2023}. Building on vision-language pretraining, RT-2 introduced the VLA paradigm by jointly fine-tuning a VLM to emit action tokens alongside language, enabling knowledge transfer from web-scale data to control \cite{zitkovich_rt-2_2023}. PaLM-E further explored tightly coupling multimodal perception and embodied task supervision within an LLM backbone \cite{driess_palm-e_2023}. More recently, open-source efforts such as OpenVLA and Octo have made large VLA-style policies and training recipes more accessible, highlighting the benefits of dataset diversity and standardized evaluation across robot embodiments \cite{kim_openvla_2024, ghosh_octo_2024, collaboration_open_2025}.

In parallel, several works investigate architectural and training choices to improve VLA reasoning and efficiency. VIMA formulates manipulation tasks as multimodal prompts and trains autoregressive policies to perform a broad set of procedurally generated tabletop tasks \cite{jiang_vima_2023}. CoT-VLA proposes explicit intermediate visual reasoning to improve temporal planning for manipulation \cite{zhao_cot-vla_2025}. On the efficiency side, SmolVLA and FLOWER target lower-cost training and deployment through compact architectures and optimized inference stacks \cite{shukor_smolvla_2025, reuss_flower_2025}, while NanoVLA studies lightweight VLA designs for resource-constrained platforms \cite{chen_nanovla_2025}. Our work is complementary: rather than introducing a new multimodal backbone, we use ASCII rendering to enable text-only LLMs to act as closed-loop controllers.

\subsection{LLM Agents for Embodied Control}
A complementary line of research uses LLMs as high-level planners or program synthesizers that call external perception and control tools, rather than learning an end-to-end visuomotor policy. Code-as-Policies demonstrates that LLMs can compose robot policy programs that invoke perception modules and low-level controllers, enabling flexible instruction following via tool use \cite{liang_code_2023}. VoxPoser leverages LLMs to infer affordances and constraints, and composes 3D value maps with a VLM to synthesize closed-loop trajectories through model-based planning \cite{huang_voxposer_2023}. Related efforts also explore LLM-driven coordination and planning in multi-robot settings using structured state representations for reasoning and control \cite{jiang_exploring_2025}.

Compared with these agentic or tool-using pipelines, we focus on a simpler closed-loop policy interface in which the model directly maps instruction and ASCII observation to a single constrained domain-specific language (DSL) action at each step. This design avoids external planners or skill libraries at inference time, while still benefiting from a planning-based teacher and DAgger-style supervision during training.

\subsection{Textualized or Symbolic Visual Representations}
Several approaches reduce visual complexity by converting observations into structured, symbolic representations that are easier for language models to process. SayPlan grounds LLM planning with 3D scene graphs, combining semantic search with classical path planning and iterative feasibility checking for long-horizon tasks \cite{rana_sayplan_2023}. GRID similarly leverages scene graphs and graph neural networks to decompose instructions into robot-relevant subtasks, emphasizing real-time inference and robustness to scene variation \cite{ni_grid_2024}. These methods highlight the value of explicit structure and abstraction for grounding and planning. ASCIIEval and ASCIIBench benchmark models' ability to recover visual structure from ASCII-formatted text~\cite{jia_asciieval_2025,luo_asciibench_2025}.

Our approach differs in the choice of representation and target setting. Rather than extracting a scene graph or symbolic state, we render the full observation into a deterministic, fixed-size, and color-aware ASCII raster that preserves spatial layout.

\section{Methodology}

\subsection{Problem Formulation}
We study a VLA control problem in which an agent must execute multi-step manipulation actions from visual observations and natural-language instructions. Each episode is defined from a top-down workspace with four types of entities: a gripper, movable objects, static obstacles, and target markers. Objects and targets are distinguished by color and shape (details in Sec.~\ref{sec:env_tasks}). The task is specified by a natural-language instruction describing a sequence of goals (e.g., visiting colored targets) and pick-and-place operations (e.g., moving a specified object to a specified target). At time step $t$, the agent receives the current visual observation $o_t$ and the instruction $I$, and produces an action $a_t$. The environment then transitions deterministically, returning a new observation $o_{t+1}$ and a termination signal.

\begin{figure*}[t]
  \centering
  \includegraphics[width=\textwidth]{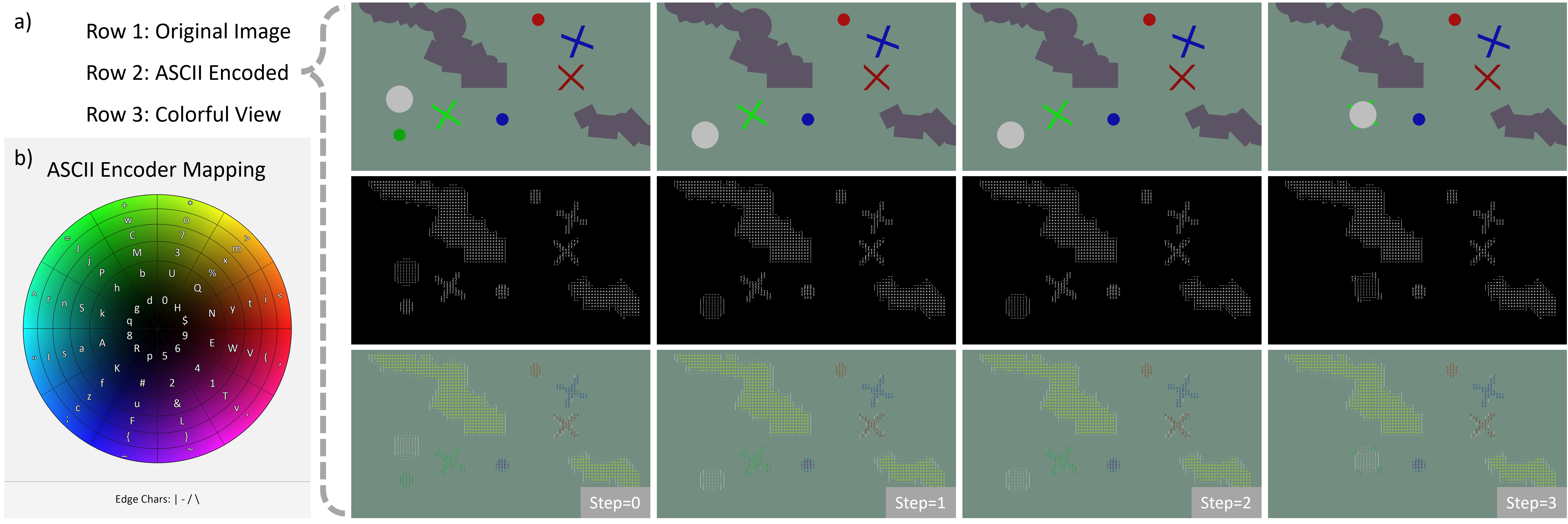}
  \caption{(a) Example of a generated \texttt{Complex} training scene (seed 26), showing steps 0--3 of a 13-step episode. The workspace contains three-colored objects and targets, obstacles, and a white circular gripper. The second row shows the ASCII encoding result. The third row was presented for human readability. (b) The ASCII encoding palette for our ASCII encoder. It is encoded in the HSV color space and has four edge characters to separate borders.}
  \label{fig:ascii_encoding}
\end{figure*}

In addition to actions, the agent also determines task feasibility. We consider three types of infeasibility, each associated with a distinct label:
\begin{itemize}
    \item \textbf{BLOCKED}: the instruction is physically infeasible because obstacles block all collision-free paths to complete at least one necessary subgoal.
    \item \textbf{NARROW}: a geometric path exists for an idealized point gripper, but not for the real gripper due to its size.
    \item \textbf{UNDEFINED}: the instruction refers to a missing entity (e.g., a requested object/target color does not exist in the scene), making the task semantically infeasible.
\end{itemize}
In unreachable episodes, the policy is expected to output \texttt{UNREACHABLE} with the appropriate subtype rather than attempt motions that will cause collisions. Episodes terminate upon (i) successful completion of the instruction, (ii) collision with an obstacle, (iii) an explicit infeasibility declaration by the agent, or (iv) reaching a fixed step budget (Sec.~\ref{sec:metrics}).

\subsection{ASCII-as-Vision Observation Interface}\label{sec:ascii_obs}

To enable LLMs to condition on visual information, we introduce an \emph{ASCII-as-vision} interface. Each RGB camera frame is rendered into a fixed-size, colored ASCII raster, which is fed to the LLM as text. The encoder is designed to preserve task-relevant visual cues, such as objects, targets, and spatial layout, while producing a deterministic, human-readable representation. Background pixels are mapped to whitespace to reduce clutter. In our experiments, we use a 96$\times$54 ASCII resolution (Sec.~\ref{sec:env_tasks}). Fig.~\ref{fig:ascii_encoding} illustrates representative simulator frames and the resulting ASCII renderings produced by our encoder, together with the fixed character palette used for color-aware discretization and border delimiters.

\begin{figure}[!t]
  \centering
  \includegraphics[width=\columnwidth]{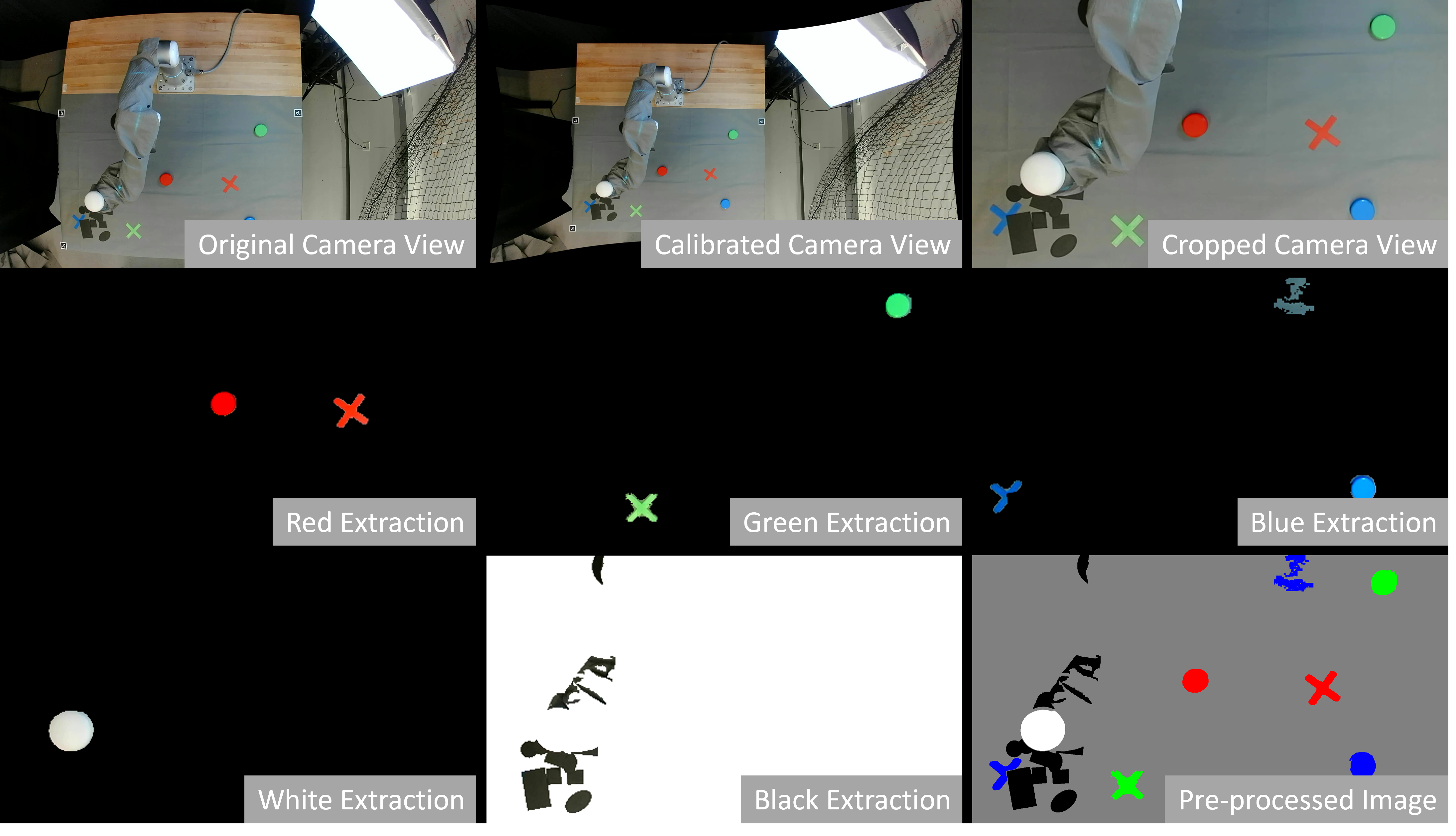}
  \caption{Real-robot image masking and denoising prior to ASCII encoding. The top row shows the raw camera input (left), the calibrated camera view (middle), and the cropped workspace view (right). The middle row visualizes color-based masks for the red/green/blue task entities before and after denoising. The bottom row shows white gripper (left), and black obstacle (middle) masks. The bottom right is the result of image pre-processing.}
  \label{fig:real_preprocess}
  \vspace{-15pt}
\end{figure}

In simulation, environments are procedurally generated using \texttt{pygame}. Each scene contains four entity types: (i) movable objects, (ii) target markers, (iii) a gripper, and (iv) static obstacles. Movable objects and targets use a canonical red/green/blue palette, while the gripper is rendered in white. Obstacles are rendered in colors outside the task palette and are chosen to maintain sufficient contrast against the background and reduce ambiguity with task-relevant entities. To improve dataset diversity and robustness of the ASCII encoder, we randomize appearance and geometry. Colors of all entities and the background vary around their nominal values, object sizes vary within a fixed interval, and both targets and obstacles receive random rotations. This domain randomization broadens the pixel values and shapes observed during training, while ensuring semantic labels.

For real-world experiments, we use a fixed overhead RGB camera and apply a deterministic pre-processing pipeline before ASCII rendering (Fig.~\ref{fig:real_preprocess}). The camera is calibrated using a checkerboard pattern. We then define a planar workspace by placing four AprilTags at the corners of the operating region. The incoming image is cropped to the AprilTag-bounded region (144\,cm $\times$ 81\,cm) and resampled to $1920\times1080$ pixels to match the simulator’s canonical aspect ratio. In the cropped image, we extract task-relevant colors via masking, followed by simple denoising to suppress spurious detections.

A key real-to-sim alignment issue is that the gripper is physically closer to the camera than the tabletop plane, causing a parallax offset to the objects and targets on the table. To correct this effect, we detect the gripper as the largest connected component in the white mask, estimate its center and radius in image coordinates, and remap only the gripper center onto the tabletop reference frame using a calibrated pixel-to-pixel mapping model fitted from collected correspondences. We then redraw a circular gripper mask at the remapped center, with the same radius, and recompose it using the denoised color masks before ASCII rendering.

\subsection{Action Space and Output Constraints}
The policy outputs a single command in a constrained DSL:
\begin{itemize}
    \item \textbf{Motion}: \texttt{MOVE(x,y)}
    \item \textbf{Grasping}: \texttt{GRIP(PICK)} and \texttt{GRIP(DROP)}
    \item \textbf{Unreachability}:\\ \texttt{UNREACHABLE(BLOCKED|NARROW|UNDEFINED)}
\end{itemize}
To simplify parsing and execution, the \texttt{MOVE} command uses coordinates on a discrete action grid rather than raw pixels. The workspace is discretized into an $W\times H$ grid, and each grid unit maps to a fixed number of pixels (Sec.~\ref{sec:env_tasks}). The model outputs exactly one valid DSL action at each step, and invalid or malformed outputs are treated as failures. % during evaluation.

\subsection{Learning Pipeline}
We train the controller in stages. In \textbf{Stage~0}, we generate supervised training data using an expert motion planner policy, A*, that has access to the environment’s geometric ground truth~\cite{4082128}. The teacher follows the instruction’s internal plan and produces step-wise actions, using planning-based motion selection and deterministic grasping rules.

From \textbf{Stage~1} onward, we employ DAgger data aggregation, in which the current policy interacts with the environment and the teacher provides corrective labels for the states visited by the policy \cite{ross_reduction_2011}. This addresses the problem of compounding errors that arise when a policy is trained only on expert trajectories but executed under its own state distribution. We adapt each base LLM with lightweight fine-tuning (LoRA) \cite{hu_lora_2021}.

\section{Experimental Setup}\label{sec:exp_setup}

\subsection{Environment and Tasks}\label{sec:env_tasks}

\begin{figure}[h]
  \centering
  \includegraphics[width=\columnwidth]{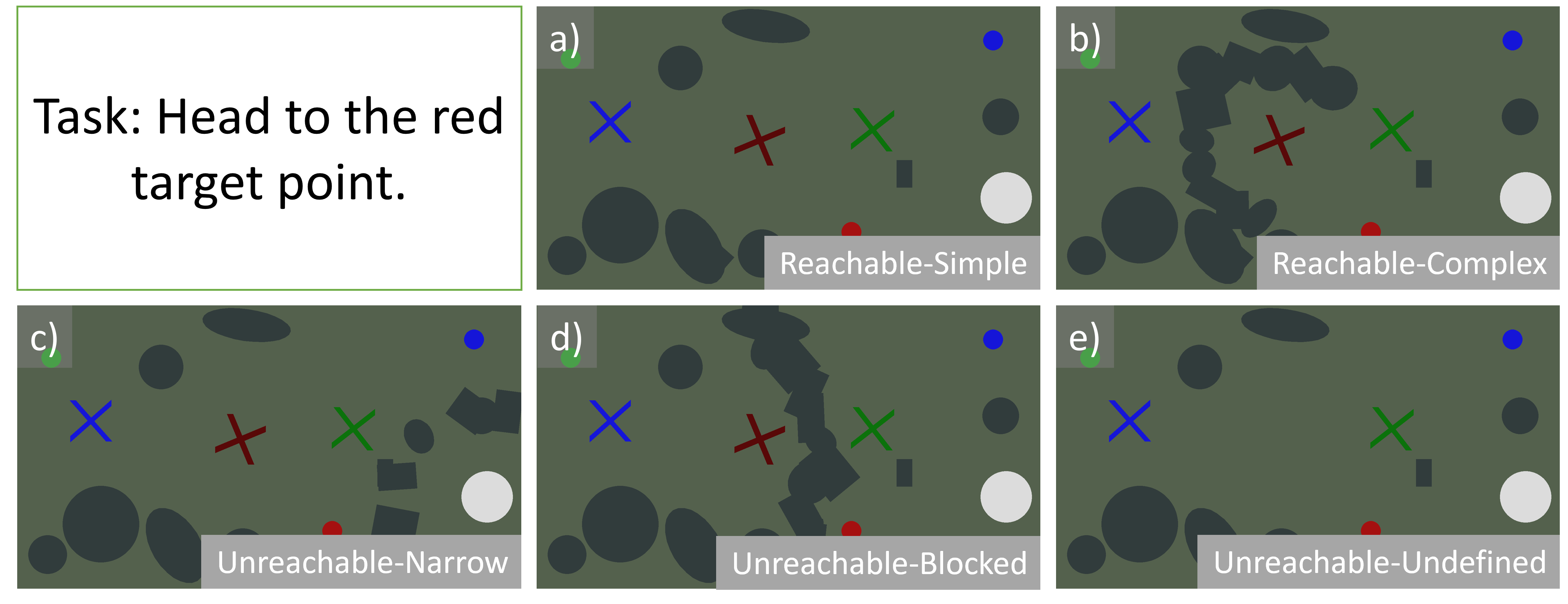}
  \caption{Reachability regimes and controlled scene generation. We first generate a \texttt{Reachable-Simple} scene by sampling a task instruction and randomly placing all required entities (gripper, graspable objects, colored targets, and obstacles) so that the task is feasible. We then compute a nominal execution path with A* and construct a \texttt{Unreachable-Blocked} variant by adding a separating obstacle that intersects the A* path and disconnects the required motion, using one of two barrier instantiations: (i) a straight barrier spanning the workspace, or (ii) a circular barrier centered at one of the path-endpoint entities, with the radius set by the distance from that entity to a randomly sampled point along the A* path. From this blocked construction, we derive the \texttt{Reachable-Complex} by removing only the path-cutting barrier while keeping the other newly introduced obstacles to increase clutter yet preserve feasibility. We derive \texttt{Unreachable-Narrow} by partially removing blocking obstacles to create gaps that exist geometrically but do not satisfy the gripper safety clearance. We derive \texttt{Unreachable-Undefined} by modifying the task or scene such that an instruction-referenced entity is missing. (a) \texttt{Reachable-Simple}. (b) \texttt{Reachable-Complex}. (c) \texttt{Unreachable-Narrow}. (d) \texttt{Unreachable-Blocked}. (e) \texttt{Unreachable-Undefined} (e.g., ``Head to the red target point'' with no red target present).}
  \label{fig:reachability}
  \vspace{-15pt}
\end{figure}

We evaluate in a 2D top-down manipulation environment implemented with \texttt{pygame} for rendering and geometric collision checks. The workspace has a 16:9 aspect ratio at 1920$\times$1080 pixels. The gripper is rendered as a large white circle; objects are rendered as smaller circles in red, green, or blue with slight color and size variation; targets are rendered as size-variant and rotated X-shaped markers in the same color set; obstacles are rendered using multiple shapes (rectangles, circles, ellipses) and colors besides red, green, or blue. Background and obstacle colors are chosen to maintain visual separability from task-relevant entities.

We evaluate five reachability regimes that separate geometric feasibility from semantic feasibility (Fig.~\ref{fig:reachability}). Starting from a \texttt{Reachable-Simple} scene, we compute a nominal A* path and create \texttt{Unreachable-Blocked} scenes by inserting a barrier that intersects the path and separates the workspace, instantiated either as a straight obstacle spanning the scene or as a circular obstacle centered at one endpoint entity with radius set by a sampled point along the path. We then derive \texttt{Reachable-Complex} by removing only the path-cutting barrier while keeping other added obstacles, preserving feasibility under increased clutter. For \texttt{Unreachable-Narrow}, we partially remove blocking obstacles to form gaps that exist geometrically but violate the gripper's clearance constraints. Finally, \texttt{Unreachable-Undefined} is created by modifying the scene or instruction so that a referenced entity is missing.

The simulation executes \texttt{MOVE(x,y)} as a straight line motion from the current gripper position to the target coordinate. If the linear path intersects any obstacle within the gripper’s radius, the episode terminates with a collision. \texttt{GRIP(PICK)} succeeds if the gripper is within a fixed grasp threshold of a movable object. \texttt{GRIP(DROP)} releases the object, and task completion is determined by whether the instruction-defined sequence of subgoals has been satisfied. Instructions are generated procedurally from templates and may contain multiple clauses. For evaluation, the environment maintains an internal plan derived from the instruction to determine success, without requiring a separate ``DONE'' action.

To make the \texttt{MOVE(x,y)} command compact and easy to parse, we discretize the continuous workspace into a fixed-resolution action grid. Concretely, for a rendered workspace of size $1920\times1080$ pixels, we use a $192\times108$ grid, i.e., each grid cell corresponds to a $10\times10$ pixel square. A discrete action $\texttt{MOVE}(x,y)$ specifies an integer grid coordinate with $x\in\{0,\ldots,191\}$ and $y\in\{0,\ldots,107\}$, which is mapped to the center of the corresponding pixel cell for execution.

\subsection{Models and Training}
We fine-tune and compare a diverse set of LLMs and VLMs, covering multiple families and parameter scales (Table~\ref{tab:model_list}). All models are adapted with LoRA using the same ASCII observation interface and the same constrained action DSL. We evaluate gemma-3-4b-it in two modes (with and without image inputs).
To ensure a controlled comparison, we standardize the effective visual resolution seen by the policy. Specifically, before feeding images to a VLM's vision encoders, we downsample the RGB observations to $96\times54$, matching the ASCII raster resolution used for LLMs. This prevents performance differences from being trivially explained by higher input image resolution in VLMs and isolates the impact of the modality interface and backbone architecture under a similar information budget.

\begin{table}[!t]
\caption{Model list used in our experiments.}
\label{tab:model_list}
\centering
\begin{tabular}{lc}
\hline
\textbf{Model} & \textbf{Modality} \\
\hline
Qwen3-0.6B & Text \\
Qwen3-1.7B & Text \\
Qwen3-4B   & Text \\
Qwen3-8B   & Text \\
Qwen3-14B  & Text \\ \hline
Qwen3-VL-2B-Instruct & VL \\
Qwen3-VL-4B-Instruct & VL \\
Qwen3-VL-8B-Instruct & VL \\ \hline
Llama-3.2-1B-Instruct & Text \\
Llama-3.2-3B-Instruct & Text \\ \hline
gemma-3-4b-it & Text \\
gemma-3-4b-it & VL \\
\hline
\end{tabular}
\end{table}

\begin{figure*}[t]
  \centering
  \includegraphics[width=\textwidth]{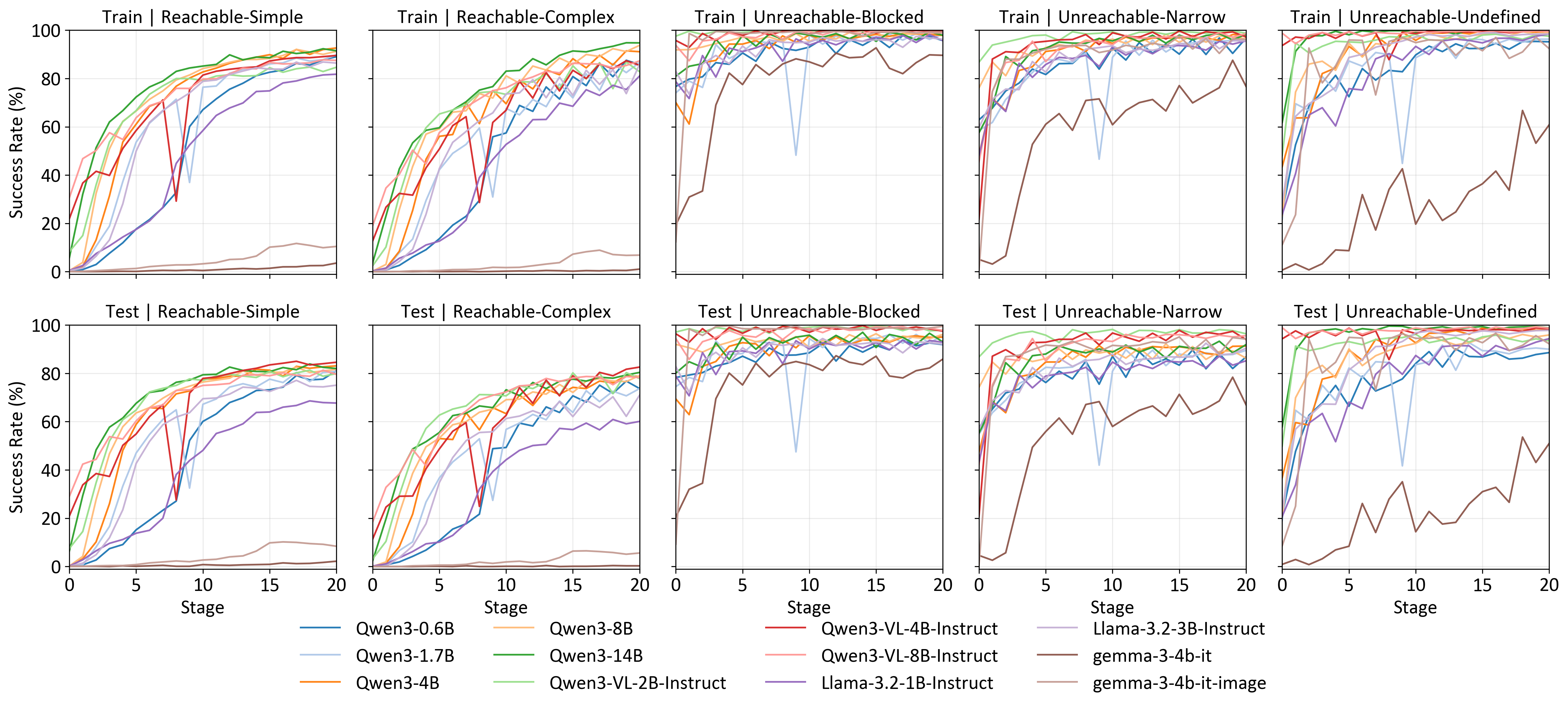}
  \caption{Stage-wise success rates in simulation. The 2$\times$5 grid reports \texttt{train}/\texttt{test} (rows) across the five scenario types (columns): \texttt{reachable-simple}, \texttt{reachable-complex}, \texttt{unreachable-blocked}, \texttt{unreachable-narrow}, and \texttt{unreachable-undefined}.}
  \label{fig:success_rate}
\end{figure*}

Training proceeds over staged datasets. Stage~0 uses teacher-generated demonstrations, while Stage~1--20 use DAgger-collected trajectories labeled by the teacher. We partition reachable episodes into two difficulty levels (\texttt{Simple} and \texttt{Complex}) and maintain a fixed per-stage data mixture where \texttt{Simple-Reachable} contributes 8000 episodes per stage, while each of the remaining regimes contributes 2000 episodes per stage, including \texttt{Complex-Reachable}, \texttt{Unreachable-Blocked}, \texttt{Unreachable-Narrow}, and \texttt{Unreachable-Undefined}. We evaluate on unseen seeds. % not used for training.

Our DAgger implementation queries the teacher at every step along learner rollouts. The model selects an action, the environment transitions to a new state, and the teacher then provides the corrective label for that visited state in a loop until termination. The teacher has access to ground truth and always returns a feasible action, providing reliable supervision. Unless otherwise noted, DAgger data accounts for 25\% of the training mixture. For Gemma models, we increase this fraction to 50\% to mitigate observed training instabilities.

For all models, we set the maximum sequence length to 4096, which accommodates the longest training examples without truncation for all sequences in our dataset. We apply LoRA to fine-tune each model.

\subsection{Metrics}\label{sec:metrics}
We report performance separately for each reachability regime. For reachable scenes, the metric is task success rate, defined as completing the full instruction without collisions and without issuing an \texttt{UNREACHABLE} action. We additionally report collision rates and average number of steps over all reachable episodes (\textbf{Steps (all)} in Table~\ref{tab:stage20_metrics}).

For unreachable scenes, we evaluate (i) unreachability detection, which is correct if the model outputs \texttt{UNREACHABLE(*)} at any step before collision or timeout, and (ii) subtype accuracy, which requires an exact match between the predicted subtype and the ground-truth label in \{\texttt{BLOCKED}, \texttt{NARROW}, \texttt{UNDEFINED}\}. We also report the false-positive rate in reachable scenes, defined as the fraction of reachable episodes in which the model outputs \texttt{UNREACHABLE(*)}.

To prevent degenerate looping behaviors, each episode has a maximum horizon (30 steps). Episodes that reach the limit without success are counted as failures.

We summarize the Stage~20 evaluation results under this metric suite in Table~\ref{tab:stage20_metrics}.

\begin{table*}[t]
\caption{Stage~20 test performance under the metrics defined in Sec.~\ref{sec:metrics}. Reported values are aggregated over held-out test data. The arrows indicate the direction of better performance in each column.}
\label{tab:stage20_metrics}
\centering
\begin{tabular}{lcccccc}
\hline
\textbf{Model} &
\textbf{Reachable} &
\textbf{Reachable} &
\textbf{Reachable} &
\textbf{Unreach.} &
\textbf{Unreach.} &
\textbf{Reachable} \\
&
\textbf{Success (\%)} $\uparrow$ &
\textbf{Collision (\%)} $\downarrow$ &
\textbf{Steps (all)} $\downarrow$ &
\textbf{Detect (\%)} $\uparrow$ &
\textbf{Subtype Acc. (\%)} $\uparrow$ &
\textbf{FP Unreach. (\%)} $\downarrow$ \\
\hline
Qwen3-0.6B & 77.4 & 13.1 & 8.43 & 94.9 & 93.9 & 5.7 \\
Qwen3-1.7B & 77.2 & 14.8 & 7.68 & 95.3 & 96.3 & 6.4 \\
Qwen3-4B & 80.9 & 13.1 & 7.72 & 97.4 & 97.3 & 5.4 \\
Qwen3-8B & 80.4 & 13.4 & 7.67 & 96.2 & 96.4 & 5.4 \\
Qwen3-14B & 81.2 & 12.3 & 7.68 & 97.8 & 96.7 & 6.2 \\
\hline
Qwen3-VL-2B-Instruct & 78.8 & 8.0 & 9.24 & 96.8 & 99.1 & 6.9 \\
Qwen3-VL-4B-Instruct & 83.6 & 8.6 & 8.51 & 97.4 & 99.2 & 6.0 \\
Qwen3-VL-8B-Instruct & 79.0 & 12.3 & 7.87 & 98.5 & 98.8 & 6.9 \\ \hline
Llama-3.2-1B-Instruct & 63.9 & 21.0 & 7.81 & 97.0 & 93.6 & 12.8 \\
Llama-3.2-3B-Instruct & 73.0 & 18.1 & 8.07 & 97.9 & 95.4 & 5.8 \\
\hline
gemma-3-4b-it (text) & 1.2 & 74.1 & 5.91 & 78.4 & 86.7 & 17.4 \\
gemma-3-4b-it (image) & 7.0 & 45.2 & 12.67 & 96.4 & 98.9 & 17.8 \\
\hline
\end{tabular}
\end{table*}

\section{Results}

\subsection{Overview of Findings}
We highlight three takeaways before presenting detailed results. First, stage-wise aggregation yields substantial gains beyond the Stage~0 bootstrap, improving closed-loop robustness across most models and scenarios (Fig.~\ref{fig:success_rate}). Second, higher-capacity backbones tend to improve both effectiveness and safety simultaneously, achieving higher reachable success while reducing collisions and false-positive \texttt{UNREACHABLE} predictions (Table~\ref{tab:stage20_metrics}). Third, under the same ASCII observation interface and action DSL, text-only LLMs and VL backbones are both competitive; at comparable scales, VL is not always superior, and stronger text-only models can approach the best VL checkpoints on several metrics (Table~\ref{tab:stage20_metrics}).

\subsection{Training Results}
Figure~\ref{fig:success_rate} shows consistent stage-wise improvement from Stage~0 to later stages across most models and scenarios. To summarize this trend, we compute a macro success score as the unweighted mean of the five scenario success rates in Fig.~\ref{fig:success_rate}. For example, on test episodes, Qwen3-4B improves from 30.8\% (Stage~0) to 89.2\% (Stage~20), Qwen3-VL-4B-Instruct improves from 48.8\% to 91.4\%, and Llama-3.2-3B-Instruct improves from 32.7\% to 85.3\%. These gains are consistent with the intended role of staged aggregation, where Stage~0 bootstraps syntax and basic planning behavior, and later stages improve closed-loop robustness under learner-visited states. For unreachable regimes, a trial is counted as successful if the policy emits \texttt{UNREACHABLE} with the correct subtype.

\subsection{Evaluation Results}
Table~\ref{tab:stage20_metrics} summarizes Stage~20 test performance under the metric suite in Sec.~\ref{sec:metrics}. The strongest overall reachable performance is achieved by Qwen3-VL-4B-Instruct (83.6\% success) with comparatively low collision rate (8.6\%). Among LLMs, Qwen3-14B reaches 81.2\% reachable success. Unreachable-scene detection is high for most later stage checkpoints (roughly mid-to-high 90\% range), and subtype accuracy is also strong (generally mid-90\% to high-90\%). At the same time, model differences remain visible in safety behavior: lower-capacity models are more collision-prone and may over-predict unreachability with more false positives (FP) reported when the scene is reachable.

Within the Qwen3 family, the gap between text-only models and VLMs variants is relatively small at comparable scales, and is not always in favor of VLMs. In particular, Qwen3-8B slightly outperforms Qwen3-VL-8B-Instruct on reachable success (80.4\% vs 79.0\%) and reachable FP-unreachability rate (5.4\% vs 6.9\%). This supports our central claim that ASCII-as-vision can substitute for native image embeddings under the same action interface and training pipeline.

\subsection{Real-World Experimental Results}

\begin{figure*}[t]
  \centering
  \includegraphics[width=\textwidth]{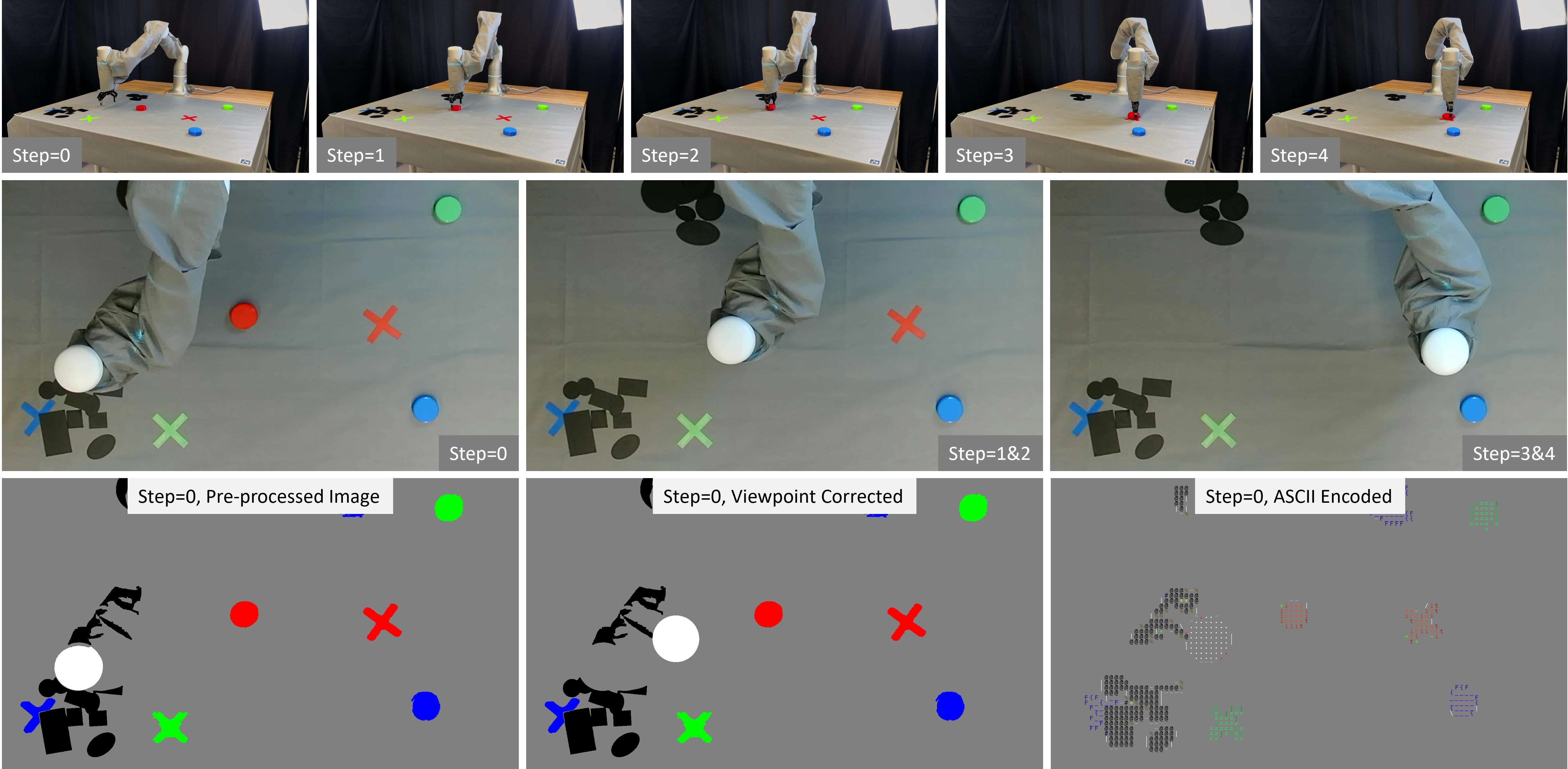}
  \caption{Real-robot pick-and-place rollout under the instruction ``Pick the red object and place it at the red target.'' The first row shows an external camera view, while the second row shows the corresponding top-down view after workspace cropping in our perception pipeline. Step~0 is the initial state; Step~1 moves the gripper above the red object; Step~2 grasps the object; Step~3 moves the gripper above the red target; and Step~4 releases the object at the target. In the top-down row, the grasp and release events induce minimal visible change at this resolution, so the corresponding frames are merged for compact visualization. The third row presents the image pre-processing, viewpoint correction, and ASCII encoding of Step~0.}
  \label{fig:real_experiment}
\end{figure*}

We deploy the learned controllers on a Flexiv Rizon4s robotic arm using the same observation and action interfaces as in simulation (Fig.~\ref{fig:pipeline}). A fixed overhead RGB camera provides a top-down view of the workspace. Each frame is processed by the deterministic pipeline described in Sec.~\ref{sec:ascii_obs}. Fig.~\ref{fig:real_preprocess} summarizes this pipeline, including calibration, workspace cropping, and mask extraction/denoising prior to ASCII encoding. The additional gripper-to-tabletop viewpoint correction is described in Sec.~\ref{sec:ascii_obs} and visualized in Fig.~\ref{fig:real_experiment} (third row). The gripper center remapping uses the same calibrated mapping model as in our robot arm controller, which was calculated from logged correspondences between observed gripper pixels and tabletop-plane reference pixels. The resulting composite image is then rendered into the same fixed-size, color-aware ASCII raster used in simulation. For VL-based ASCII art VLAs, we feed the corresponding pre-processed RGB observation to the native vision encoder after downsampling to the same $96\times54$ resolution.

At each control step, the policy receives the natural-language instruction and the current ASCII observation and outputs a DSL action. \texttt{MOVE(x,y)} selects a discrete action-grid coordinate that is mapped to a Cartesian coordinate in the calibrated real workspace, and a Cartesian planner from MoveIt \cite{coleman_reducing_2014} calculates a trajectory to the waypoint. \texttt{GRIP(PICK)} and \texttt{GRIP(DROP)} actuate the gripper to grasp or release an object. The system runs in closed loop until the instruction is completed, the policy declares infeasibility via \texttt{UNREACHABLE}, or the fixed horizon is reached (Sec.~\ref{sec:metrics}).

Fig.~\ref{fig:real_experiment} shows a representative robot arm pick-and-place rollout for the instruction, ``Pick the red object and place it at the red target.'' The policy produces a multi-step sequence that moves to the object, grasps, transports, and releases at the target. Qualitatively, these demonstrations indicate that controllers trained with the ASCII-as-vision interface can transfer to hardware with minimal changes to the policy interface, with the additional requirements being deterministic perception alignment and a low-level robot controller. In practice, remaining errors are dominated by real-world perception and execution uncertainties absent from the idealized 2D simulator (e.g., lighting-induced mask noise, residual alignment error, and grasp/execution variability), suggesting clear directions for improving robustness while retaining the simplicity and interpretability of the ASCII observation channel.

\section{Discussion and Limitations}
Our policy effectively performs local geometric reasoning by selecting collision-free waypoint sequences from a low-resolution ASCII observation (\(96\times54\)). We do not aim to solve the full spectrum of robot motion planning---e.g., globally optimal planning in high-dimensional configuration spaces with dynamics and contact---but rather to test whether a standard LLM can serve as a closed-loop local planner when vision is provided via text. The remaining performance gap on \texttt{reachable-complex} and \texttt{unreachable-narrow} regimes suggests that tighter geometric constraints and long action sequences remain challenging, and motivates future work on more action primitives, higher-resolution (or multi-scale) ASCII encodings, and hybridization with classical planners.

Our current study has several limitations. First, ASCII rendering is an intentional information bottleneck: while it offers compatibility and interpretability, it discards fine-grained visual cues and may limit performance in cluttered scenes, small-object interactions, and tasks requiring precise geometry. Second, despite stable trends for Qwen3 and Llama-3.2 families, optimization results are model-dependent. For example, Gemma models are unstable under the same hyperparameters used for Qwen3/Llama-3.2; lowering the learning rate and adjusting the DAgger data ratio improves stability, but final performance still lags behind Qwen3 and Llama-3.2. Understanding these training discrepancies is an important direction for future work. We also observed that \texttt{reachable-complex} remains consistently harder than \texttt{reachable-simple}, and \texttt{unreachable-narrow} is often the most challenging unreachable subtype, suggesting that geometric clearance reasoning is still a key bottleneck.

Another limitation is the modality scope: we currently encode only RGB observations. A promising extension is ASCII-based encoding for RGBD inputs. Depth is often scarcer in foundation model pretraining and is less consistently supported by multimodal embedding stacks than RGB. A text interface that serializes both RGB and depth may provide a low-cost path to incorporate depth information without requiring dedicated RGBD components.

\section{Conclusion}
We presented a lightweight ASCII-as-vision interface that enables LLMs to operate as closed-loop VLAs by inputting ASCII renderings of visual inputs and outputting actions. Using a motion planner teacher and DAgger, our controllers learn multi-step manipulation behaviors in a 2D environment, with obstacle avoidance, pick-and-place, and abstention when an instruction is infeasible. We find that LLMs can achieve competitive performance with the same action interface and training pipeline, and, in some cases, match or exceed similarly sized VLMs. We further demonstrated transfer to a real robot arm by applying a deterministic pre-processing pipeline to align real camera observations with the simulator’s viewpoint and then using the same ASCII encoder and action DSL at inference. Overall, these results suggest that ASCII rendering can serve as a practical, interpretable modality bridge that lowers engineering barriers for VLA research and motivates future work on more diverse sensing modalities, environments, and robustness under visual and geometric variation.

\section*{Supplementary Materials}
% Hidden due to blind review, will present afterward.

Models and Result Data: \url{https://huggingface.co/cccat6/ASCII_Art_VLA/tree/main}

Training Dataset: \url{https://huggingface.co/datasets/cccat6/ASCII_Art_VLA/tree/main}

\section*{Acknowledgment}
We utilized LLM tools, including Codex, Claude, and GPT, which were used to assist with software implementation, writing, proofreading, and literature search. All final text and references were edited and reviewed by humans. RX is supported by the National Science Foundation under Grant DGE-\#\#\#\#\#\#\#. Any opinions, findings, conclusions, or recommendations expressed in this material are those of the authors and may not reflect those of the NSF.
% Full Grant Number: DGE-2236868

\bstctlcite{IEEEexample:BSTcontrol}
\bibliographystyle{IEEEtranS}
% \bibliographystyle{IEEEtran}
% \bibliography{references}
\bibliography{refs}

\end{document}